\documentclass{article}
\usepackage{spconf,amsmath,graphicx}
\usepackage{multirow}
\usepackage{array}
\usepackage{boldline}
\usepackage{enumitem}
\usepackage[dvipsnames]{xcolor}
\usepackage[lofdepth,lotdepth]{subfig}

\usepackage{cite}
\title{Dialogue history integration into end-to-end signal-to-concept spoken language understanding systems}
\name{Natalia Tomashenko$^1$, Christian Raymond$^2$, Antoine Caubri\`ere$^{3,1}$, Renato De Mori$^{1,4}$, Yannick Est\`eve$^1$\thanks{{This work was supported by the French ANR Agency through
the ONTRAC and AISSPER projects, under the contracts ANR-18-
CE23-0021-01 and ANR-19-CE23-0004-01, and by the RFI Atlanstic2020 RAPACE
project.}}}
\address{$^1$LIA - Avignon Universit\'e - France\\
$^2$INSA Rennes/IRISA - Rennes, France\\
$^3$LIUM - Le Mans Universit\'e - France\\
$^4$McGill University - Montreal, Qu\'ebec, Canada
}
\begin{document}
\ninept
\maketitle
\begin{abstract}
This work investigates the embeddings for representing dialog history in spoken language understanding (SLU) systems.
We focus on the scenario when the semantic information is extracted directly from the speech signal by means of a single end-to-end neural network model.
We proposed to integrate dialogue history into an end-to-end signal-to-concept SLU system.
The dialog history is represented in the form of dialog history embedding vectors (so-called \textit{h-vectors}) and is provided as an additional information  to  end-to-end SLU models in order to improve the system performance.
Three following types of h-vectors are proposed and experimentally evaluated in this paper: (1) \textit{supervised-all} embeddings predicting bag-of-concepts expected in the  answer of the  user from the last dialog  system response; (2) \textit{supervised-freq} embeddings focusing on predicting  only a selected set of semantic concept (corresponding to the most frequent errors in our experiments); and (3) \textit{unsupervised} embeddings. 
Experiments on the MEDIA corpus for the semantic slot filling task demonstrate that the proposed h-vectors improve the  model performance.

\end{abstract}
\section{Introduction}
\label{sec:intro}

The task of spoken language understanding (SLU) system is to detect fragments of semantic knowledge in speech data.
Popular models are made of frames describing relations between entities and their properties~\cite{tur2011spoken,shen2019modeling,li2019incremental}.
The SLU system instantiates a predefined set of frame structures called concepts that can
be mentioned in a sentence or a dialogue turn. Concept mentions express dialogue acts~(DA), intents, domain knowledge, and frame properties often represented by slots, identified by entity names, and slot filler values identified by mention types.
Concept mentions are difficult to characterize in terms of words or characters. They may be localized by head words or short word sequences called concept supports. 
For example, word spans can be hypothesized to be mentions of concepts, while entire sentence can be considered for hypothesizing dialogue acts. Unfortunately, mentions may be ambiguous because their word spans may express more semantic constituents, be incomplete or be affected by errors of an automatic speech recognition (ASR) system. 
These difficulties can be alleviated by considering certain head words, word spans, or a sentence as a seed for hypotheses generation and using additional context for providing predictions useful for constraining instantiation decision. 
An example of additional distant context used so far is a representation of dialogue history made of embeddings of sentences preceding the sentence or dialogue turn to be interpreted~\cite{chen2016end, goo2018slot, zhao2019hierarchical, sankar2019neural, goel2019hyst,history,henaff2016tracking,korpusik2019dialogue,lee2019sumbt}.

A problem that has not yet thoroughly investigated is to select what to embed and how. Some popular corpora used so far (e.g. ATIS~\cite{dahl1994expanding}) do not have explicit sentence history. In this case, the only context to pay attention to is the sentence to be interpreted. If some history information is available, then distant contexts for DA and concepts may be different.
Specific contexts for DA have been proposed in~\cite{liu2017using, ortega2019context}. For concepts, the selection of  distant contexts may depend on the complexity of the application semantic domain. For example, the French MEDIA corpus~\cite{devillers2004french} has concepts of reference, relative time, locations, prices, logical conjunction and disjunction that are expressed by short semantically ambiguous words, which are often difficult to recognize, requiring knowledge of a semantic context called state of-the world to reduce the perplexity.
Furthermore, the problem of  deciding the type of embedding is also relevant as made evident in recent published papers~\cite{komninos2016dependency, 
lin2017structured, peters2018deep, 
yin2018dimensionality, zhang2018diffusion, yaghoobzadeh2019probing}.

 In this paper, we investigate the use of different types of dialog history representation, extracted with or without supervision, and their impact on the performance of an end-to-end signal-to-concept neural network.

Noticeable approaches for reducing uncertainty in concept detection automatically extract relevant information from dialogue history~\cite{chen2016end,goo2018slot,goel2019hyst}. Considering the concern expressed in [7] and prior knowledge, we propose to focus on types of history contents starting by considering the previous system turn that contains semantically unambiguous information. In fact, the sequence of words in the system turn is generated by a semantic model whose goal is to reach a commit state for performing a transaction. Furthermore, using the train set, it is possible to compute prediction probabilities of user enunciated concepts, given the system enunciated concepts. The most likely predicted concepts can thus be used for reducing interpretation uncertainty in the following user turn.

The rest of the paper is organized as follows.
Section~\ref{sec:architecture} presents an architecture of an end-to-end  signal-to-concept model  and the proposed way of integration of dialog history ebmeddings (to which we refer as  \textit{h-vectors})  into this model.
Section~\ref{sec:dialrep} introduces different ways to represent the dialog history.
Sections~\ref{sec:exp} describes the experimental setup and results. Finally, the conclusions are given in Section~\ref{sec:concl}.

\section{End-to-end signal-to-concept neural architecture}
\label{sec:architecture}

\begin{figure}[tp]
\centering\includegraphics[width=80mm]{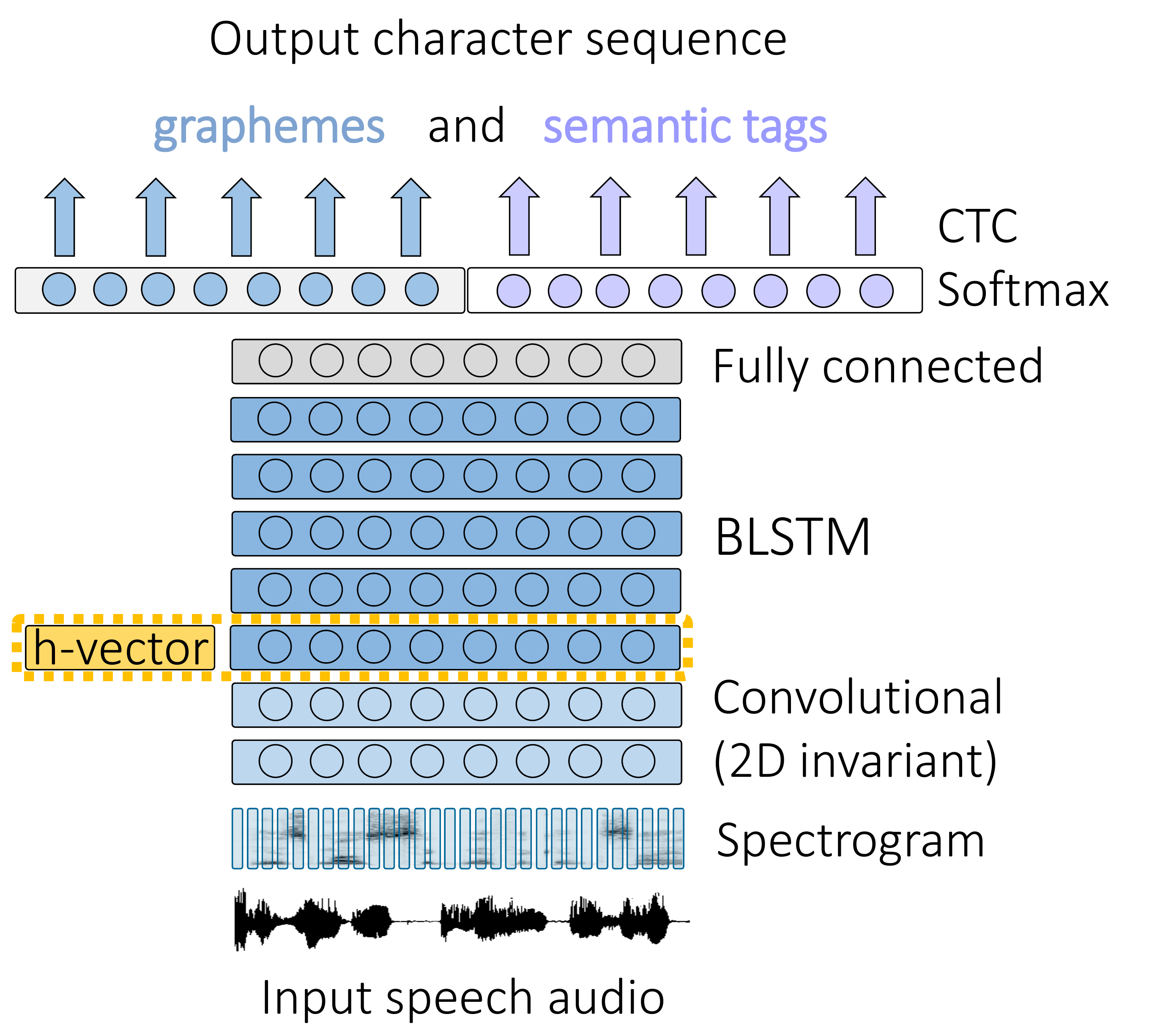}
\caption{End-to-end concept-to-semantic deep neural network model  architecture. H-vectors represent dialog history embeddings vectors.}
\label{fig:ds}
\end{figure}

Nowadays there is a growing research interest  in end-to-end systems for various SLU tasks~\cite{qian2017exploring,haghani2018audio,serdyuk2018towards,ghannay2018end,chen2018spoken,lugosch2019speech,tomashenko2019investigating,tomashenko2019recent,caubriere2019curriculum}.
In this work, similarly to~\cite{ghannay2018end,tomashenko2019investigating}, end-to-end training of signal-to-concept models is performed through the recurrent neural network (RNN) architecture and the connectionist temporal classification (CTC) loss function~\cite{graves2006connectionist} as shown in Figure~\ref{fig:ds}.
A spectrogram of power normalized audio clips  calculated on 20ms windows is used as the input features for the system.
As shown in Figure~\ref{fig:ds}, it is followed by 2D-invariant (in the time and-frequency domain) convolutional layers, and then  BLSTM layers.
A fully connected layer is applied after BLSTM layers, and  the output layer of the neural network  is a softmax layer.  
The model is trained using the CTC loss function.
H-vectors are appended to the outputs of the last (second) convolutional  layer, just before the first recurrent (BLSTM) layer.

The outputs of the network  consist of the two subsets: (1) outputs to represent the words (graphemes of a corresponding language, a \textit{space} symbol to denote word boundaries, and a \textit{blank} symbol), and (2) outputs to represent semantic concepts types and a closing symbol for semantic tags. 
We have several symbols corresponding to semantic concepts (in
the text these characters are situated before the beginning of a
semantic concept, which can be a single word or a sequence of several words) and a one tag corresponding to the end of the semantic concept, which is the same for all semantic concepts.

In order to improve model performance, we integrate dialog history information in  form of h-vectors into the model as shown in Figure~\ref{fig:ds}.
Each h-vector is calculated from the last dialog system response as described further in Section~\ref{sec:dialrep}.

H-vectors are appended to the outputs of the last (second) convolutional  layer, just before the first recurrent (BLSTM) layer.
In this paper, 
for better initialization, we first train a model using \textit{zero vectors} of the same dimension (all values are equal to~0) instead of h-vectors. Then, we use this pretrained model and finetune it on the same data but with the real h-vectors.
This approach was inspired by~\cite{deena2017semi}, where the idea of using zero auxiliary features during pretraining was implemented for language models, and by~\cite{tomashenko2019investigating}, where it was used for i-vectors.
In our preliminary experiments this type of pretraing demonstrated better results than direct model training with h-vectors, hence we use it in the experiments presented in this paper.

\section{Dialog history representation}\label{sec:dialrep}

The MEDIA corpus is a French corpus of spoken human/machine dialogues dedicated to hotel booking~\cite{devillers2004french}. 
Recently, it has been shown that this corpus is one of the current most challenging corpora for slot filling (SF) task~\cite{benchmarking} due to its complexity.
In this dataset, a human/machine dialogue is composed of 15 utterances from the user on average, and the same number from the system.

\begin{figure*}[htbp]
  \begin{center}
    \subfloat[Supervised]{
      \includegraphics[width=0.5\textwidth]{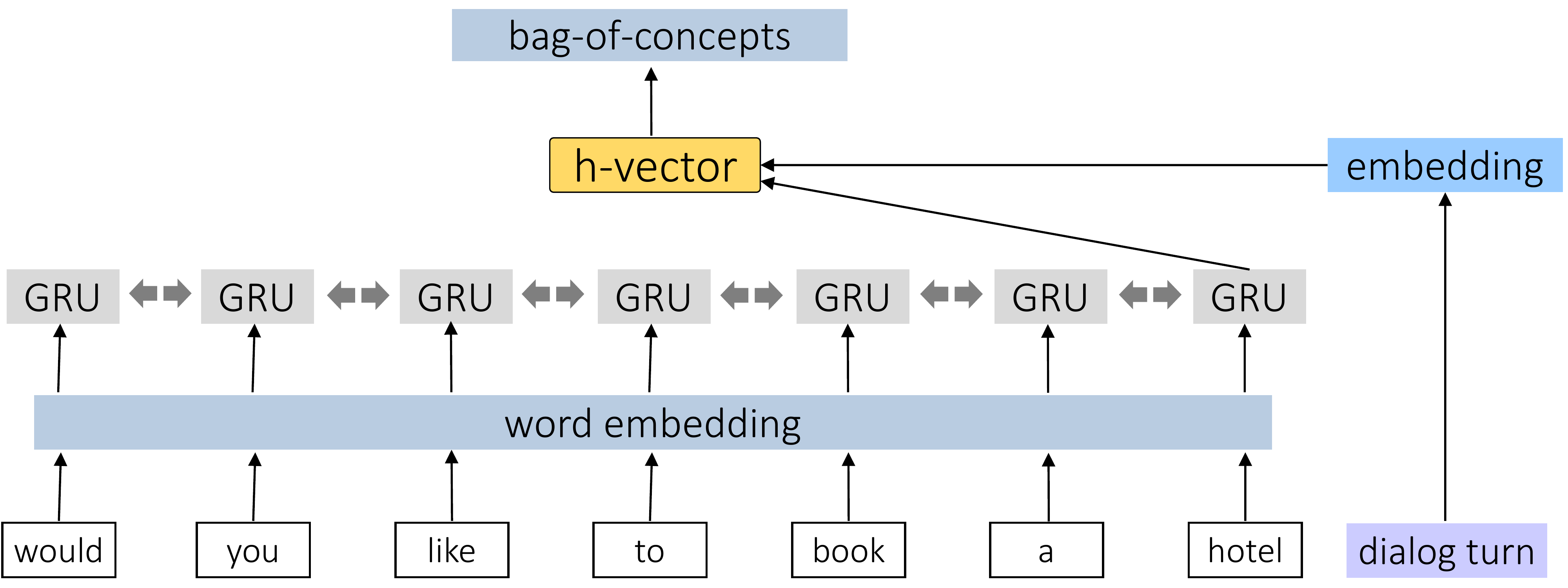}
      \label{fig:supervised}
                         }
    \subfloat[Unsupervised (autoencoder)]{
      \includegraphics[width=0.5\textwidth]{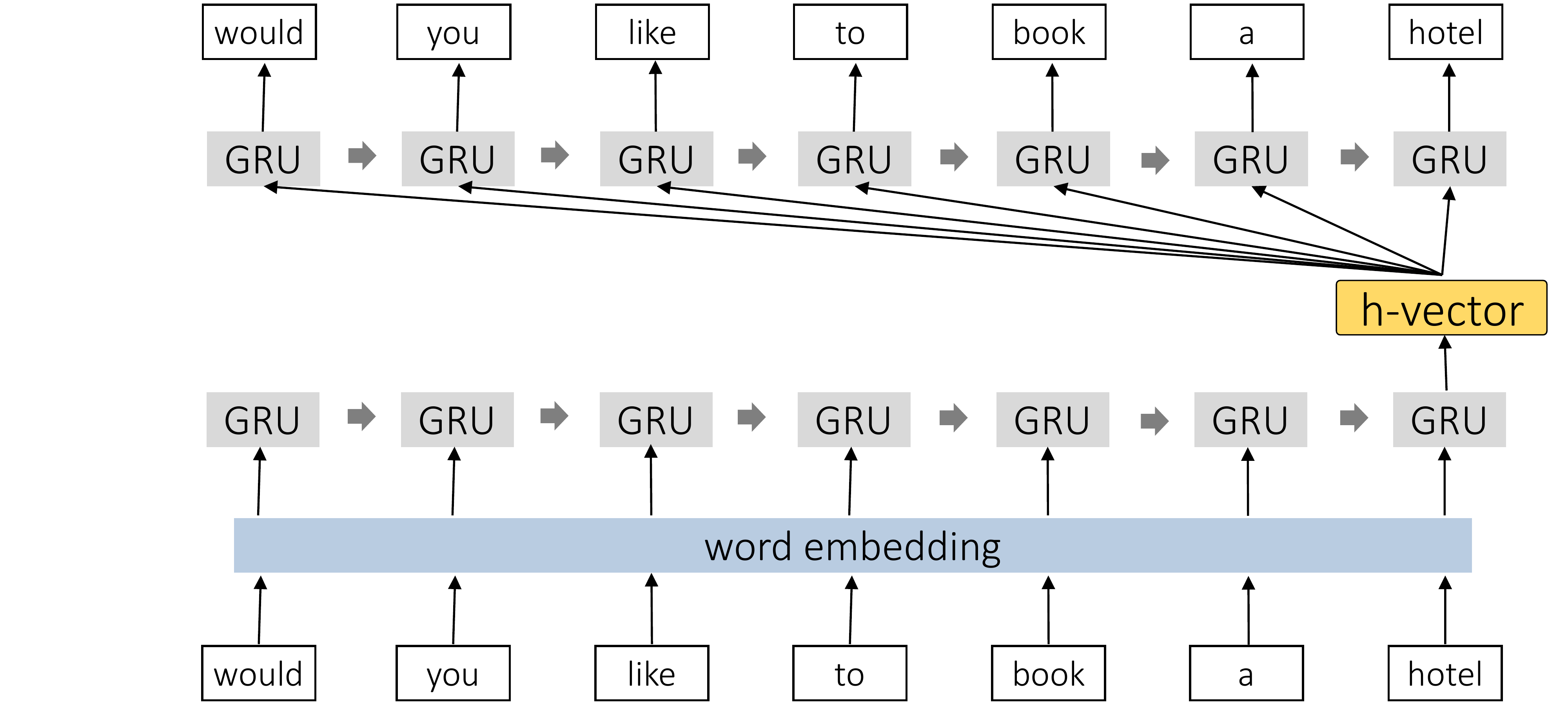}
      \label{fig:autol}
                         }
    \caption{Supervised and unsupervised  architectures for h-vector extraction}
    \label{fig:renonculacees}
  \end{center}
\end{figure*}

For this work, we decided to use as history information, the previous system prompt as it provides most of the time a good evidence of what the user answers. 
The goal is to help the main system to predict concept tags, hence  our aim is to encode the previous system prompt into an embedding that contains useful information to achieve this objective.

\subsection{Embedding with supervision}\label{sec:emb_superv}
A first h-vector type is produced using a bidirectional gated recurrent unit (GRU \cite{gru}) network to analyse the system prompt and produce a vector of embedding that is the input to a decision layer whose objective is to predict the bag of concepts of the future user answer, illustrated in Figure~\ref{fig:supervised}. The bag of concepts is represented by a vector whose size is the number of unique concepts (slots) in the application, the concepts that appears in the next user intervention are set to one. The output layer is thus a multiclass multi-output sigmoid layer and the network is trained using a binary cross-entropy loss.  As the turn in the dialog itself may identify some useful statistics, a very short part (2\%) of the h-vector is reserved to encode the dialog turn itself. Obviously, predicting the presence or absence of all the concepts of the next user answer from the previous system prompt is not possible and the network may overfit. 

\subsection{Embedding with no supervision}\label{sec:emb_unsuperv}
Another solution that is more straightforward to train is  to use a recurrent autoencoder to encode the prompt into a single h-vector. This h-vector is obtained by a symmetric neural network using a forward GRU in the encoder and decoder part, the output is a softmax layer (size is the vocabulary of the system) whose objective is to reconstruct the input prompt, illustrated in Figure~\ref{fig:autol}.

\section{Experiments}
\label{sec:exp}

\subsection{Data}
\label{ssec:data}

\vspace{-2pt}

Several publicly available corpora have been used for experiments (see Table~\ref{tab:data}).
\vspace{2pt}
\begin{table}[tbh]
  \caption{Corpus statistics  for ASR and SF tasks. }\label{tab:data}
  \centering
  \begin{tabular}{ | l | l  |l   |}
   \hline
    \multicolumn{1}{|l|}{\textbf{Task}} & \textbf{Corpora} &  \multicolumn{1}{c|}{\textbf{ Size, hours}}   \\
    \hline \hline
  {ASR train~~~ }                 & EPAC~\cite{esteve2010epac}, ESTER~1,2~\cite{galliano2009ester} & 404.6         \\
                    & ETAPE~\cite{gravier2012etape},  REPERE~\cite{giraudel2012repere} &   \\
    &                 DECODA~\cite{bechet2012decoda}, MEDIA~\cite{devillers2004french} &        \\
                          & PORTMEDIA~\cite{lefevre2012robustesse}  &     \\ \hline
                        
   SF  train                   & MEDIA (train) & 15.8        \\ \cline{2-3}
                        
    SF   dev               & MEDIA (dev) & 1.6     \\ \cline{2-3}
    SF   test                & MEDIA (test) & 4.6    \\    \hline
  \end{tabular}
  \vspace{-10pt}
\end{table}

\subsubsection{ASR data}\label{sec:asr_data}

\vspace{-4pt}
In this paper, the ASR data (audio speech files with text transcriptions) are used for  transfer learning as described in Section~\ref{sec:models}.
The corpus for ASR training is composed of  corpora from
various evaluation campaigns in the field of automatic speech processing for
French, as shown in Table~\ref{tab:data}.
The EPAC~\cite{esteve2010epac}, ESTER~1,2~\cite{galliano2009ester}, ETAPE~\cite{gravier2012etape},  REPERE~\cite{giraudel2012repere} contain transcribed speech in French from TV and  radio broadcasts.
These data were originally in the microphone channel and for experiments in this paper were  downsampled from 16kHz to 8kHz, since the test set for our main target task (SF) consists of  telephone conversations. The DECODA~\cite{bechet2012decoda} corpus is composed of  dialogues from the  call-center 
of the Paris transport authority. The  MEDIA~\cite{devillers2004french,bonneau2006results} and PORTMEDIA~\cite{lefevre2012robustesse}  are corpora of dialogues
simulating a vocal tourist information server.

\subsubsection{SF data}\label{sec:sf_data}

The  MEDIA   French corpus, dedicated to semantic extraction from speech in a context of human/machine dialogues, is used in the current experiments (see Table~\ref{tab:data}). The corpus has  manual transcription and conceptual annotation of dialogues from 250 speakers. 
It is split into the following three parts~\cite{vukotic2015time}: (1) the training set (720 dialogues, 12K sentences), (2) the development set (79 dialogues, 1.3K sentences, and (3) the test set (200 dialogues, 3K sentences).
A concept is defined by a label and a value, for example with the concept \textit{date}, the value \textit{2001/02/03} can be associated~\cite{vukotic2015time,devillers2004french}.
The MEDIA corpus is related to the hotel booking domain, and its annotation contains $76$ semantic concept tags: \textit{ room number, hotel name, location, date, room equipment}, etc. 

\subsection{H-vector extraction}\label{sec:h_exp}
  
We produced three different types of h-vectors:
two types of h-vectors using the neural architecture trained in a supervised way to predict the bag of MEDIA concepts:
\begin{itemize}
\item \textit{supervised-all  h-vectors}.
To extract these h-vectors, we trained a model as described in Section~\ref{sec:emb_superv}.  The accuracy of the model  to predict the next bag of concepts is 45\% on the train and 26\% on the test dataset.  The model has 30.382 parameters.
 \item \textit{supervised-freq h-vectors}.
This version has been trained with a bag of the four history concepts that have been observed in the train and development set to predict concepts that are frequently misrecognized.

This version tends to overfit with around 60\% of accuracy on the train and only 16\% on the test. The model has 23.918 parameters.
\end{itemize}

The third type of embeddings is trained in an unsupervised way:
\begin{itemize}
\item
\textit{unsupervised h-vectors}. These h-vectors are produced by the autoencoder architecture as described in Section~\ref{sec:emb_unsuperv}. The autoencoder  has 246.270 parameters, and the accuracy in the reconstruction is 52\% on the train and 48\% on the test.
\end{itemize}

Jointly trained word embedding is of size 10 while the dimension of h-vectors equals to  100 in all experiments.
The described architectures for h-vectors were implemented using the \textit{Keras} framework~\cite{keras}.

\subsection{Signal-to-concept models}\label{sec:models}

The neural architecture is  inspired  by  the \textit{Deep Speech~2}~\cite{amodei2016deep} for ASR.
The two major differences in comparison with the original architecture are the following. 
First, we integrated dialog history into this system based on dialog history embedding vectors (\textit{h-vectors}) as shown in Figure~\ref{fig:ds} and proposed in Section~\ref{sec:dialrep}.
Second, in this paper, the  task is SF, therefore  the output sequence besides the alphabetic characters also contains special characters corresponding to the semantic  tags~\cite{ghannay2018end, tomashenko2019investigating}.

A spectrogram of power normalized audio clips  calculated on 20ms windows is used as the input features for the system.
As shown in Figure~\ref{fig:ds}, input features are spectrograms. 
They are followed by two 2D-invariant (in the time and-frequency domain) convolutional layers\footnote{With parameters: kernel size=(41, 11), stride=(2, 2), padding=(20, 5)}, and then by five 800-dimensional BLSTM layers with sequence-wise batch normalization.   A fully connected layer is applied after BLSTM layers,  and the output layer of the neural network is a softmax layer.
The model is trained using the CTC loss function~\cite{graves2006connectionist}.
We used the \textit{deepspeech.torch} implementation\footnote{https://github.com/SeanNaren/deepspeech.pytorch}  for training baseline models, and our modification of this implementation to integrate dialog history embedding vectors.

In this work, we performed experiments with two types of models: (1) models that are trained directly on the target task using the MEDIA corpus dataset and (2) models that are trained using the transfer learning paradigm.
Transfer learning is performed from the ASR task as described  in~\cite{tomashenko2019investigating}.

For transfer learning experiments, we first trained an ASR model on the ASR data (described in Section~\ref{sec:asr_data}) using a similar end-to-end model architecture as we used for the SLU model.
The difference is in the text data preparation and output targets.
For training ASR systems, the output targets correspond to alphabetic characters and a \textit{blank}  symbol, while for slot filling task, we used additional targets corresponding to the semantic concept tags and one tag corresponding to the end of a concept.
Then, we changed the softmax layer in this model by replacing the targets with the SF targets and continue  training on the corpus annotated with semantic tags (Section~\ref{sec:sf_data}).

\subsection{Results}\label{sec:res}

Performance was evaluated in terms of   \textit{concept error rate} (CER)\footnote{CER is defined as the ratio of
the total number  of deleted, inserted and confused concepts and the total number of concepts in reference utterances.} and \textit{concept value error rate} (CVER)\footnote{CVER, in comparison to CER, takes into account  concept/value pairs instead of only concepts.} on the MEDIA test dataset.

In the first series of experiments, we trained a 
 baseline model and models with different types of h-vectors described in Section~\ref{sec:h_exp}.
 Results for these models are given in Table~\ref{tab:res1}. 
All the models in this table are trained  directly on the MEDIA training corpus. 
The first line  shows the baseline result for the end-to-end signal-to-concept model.
The other three lines (\#2,3,4) correspond to the models trained with dialog history integration and differ from each other in the way the dialog history is represented in the form of h-vectors.
We can observe, that all types of h-vectors provide an improvement over the baseline model for both metrics CER and CVER.
The best result (line \#4) is obtained for \textit{supervised-all} h-vectors and corresponds to 12.5\% of relative CER reduction and to 11.9\% of CVER reduction in comparison with the baseline model.

\begin{table}[htbp]
\caption{Slot filling performance results  on the MEDIA test dataset for the baseline model and models trained with  different types of dialog history embedding vectors. Results are given in terms of   CER and CVER metrics (\%); $\Delta$CER and $\Delta$CVER (\%) denote relative error reduction for CER and  CVER correspondingly in comparison with the baseline model (\#1). }\label{tab:res1}
\vspace{2pt}
\renewcommand{\tabcolsep}{0.09cm} 
  \centering
  \begin{tabular}{ |  l | l | l l | l l |}
  \hline
\textbf{\#~~} & 	\textbf{\shortstack{ h-vector type }} &	\textbf{CER~}	& \textbf{$\Delta$CER~}	& \textbf{CVER~}	& \textbf{$\Delta$CVER~} \\ \hline \hline
1 &	no  (baseline) &	39.2 &	- &	53.0	& - \\ \hline
2 &	unsupervised	& 35.8 & 8.7 & 47.6 & 10.2 \\
3 &	supervised-freq &	35.9 &	8.4 &	48.2 &	9.1 \\
4 &	supervised-all &	\textbf{34.3} &	\textbf{12.5} &	\textbf{46.7} &	\textbf{11.9} \\ \hline
  \end{tabular}
\end{table}

It was shown in~\cite{tomashenko2019investigating}, that transfer learning can significantly  improve the performance  of end-to-end SLU models.  
In this work, we  are also interested in exploring the proposed approach for  more accurate models trained using the transfer learning paradigm.
For this purpose, we trained two models using transfer learning from the ASR task as proposed in~\cite{tomashenko2019investigating} and described in Section~\ref{sec:models}.
Results for these models are presented in Table~\ref{tab:res2}.
The first line corresponds to a baseline model. 
The second line demonstrates the result for the model trained with the best type of dialog history embedding  vectors  (\textit{supervised-all}) chosen according to our first series of experiments.
We can see that h-vectors continue to provide an improvement  in performance over the stronger baseline: 7.7\% of relative CER reduction and 6.3\% of relative CVER reduction.

\begin{table}[tbh]
\caption{Slot filling performance results  on the MEDIA test dataset for the baseline model and the best model trained with  \textit{supervised-dall} type of dialog history embedding vectors. Models for SF are trained using \textbf{transfer learning} from an ASR model.}
\label{tab:res2}
\vspace{2pt}
\renewcommand{\tabcolsep}{0.09cm} 
  \centering
  \begin{tabular}{ |  l | c | l l | l l |}
  \hline 
\textbf{\#~~} & 	\textbf{\shortstack{ h-vector type }} &	\textbf{CER~}	& \textbf{$\Delta$CER~}	& \textbf{CVER~}	& \textbf{$\Delta$CVER~} \\ \hline \hline
1 &	no  (baseline) &	23.5 &	- &	30.0	& - \\ \hline
2 &	supervised-all &	\textbf{21.7} &	\textbf{7.7} &	\textbf{28.1} &	\textbf{6.3} \\ \hline
  \end{tabular}
\end{table}

\section{Conclusions}\label{sec:concl}
In this paper, we have proposed a novel way of integration of the dialog history information into end-to-end signal-to-concept SLU models by means of using so-called \textit{h-vectors}. 
We have proposed different types of h-vectors 
and investigated their effectiveness for end-to-end SLU using as an example the semantic slot filling task.
Experiments on the MEDIA corpus 
demonstrated that using h-vectors  improves the slot filling model performance by about 8–13\% of relative CER reduction, and by about 6-12\% of relative CVER reduction.
The best result was obtained using \textit{supervised-all} h-vectors predicting bag-of-concepts representations  of the user's answer from the last system response.

\bibliographystyle{IEEEbib-abbrev}

\ninept
\bibliography{refs}
\end{document}